%% file: main.tex
\documentclass[10pt,twocolumn,letterpaper]{article}

\usepackage{cvpr}              
\usepackage{amsmath}
\usepackage{amssymb}
\usepackage{booktabs}
\usepackage[misc]{ifsym}
\usepackage{multirow}
\usepackage{subcaption}
\usepackage{tabularx}
\usepackage[accsupp]{axessibility} 
\newcolumntype{Y}{>{\centering\arraybackslash}X}

\usepackage[pagebackref,breaklinks,colorlinks,citecolor=cvprblue]{hyperref}

\def\paperID{8490} 
\def\confName{CVPR}
\def\confYear{2024}

\title{Blur-aware Spatio-temporal Sparse Transformer for Video Deblurring}

\author{%
Huicong Zhang\textsuperscript{\rm 1}, Haozhe Xie\textsuperscript{\rm 2}, Hongxun Yao \textsuperscript{\rm 1} \textsuperscript{\Letter}\\
\textsuperscript{\rm 1} Harbin Institute of Technology 
\hspace{4 mm}
\textsuperscript{\rm 2} S-Lab, Nanyang Technological University\\
{\tt\small \url{https://vilab.hit.edu.cn/projects/bsstnet}}}

\begin{document}
\definecolor{cvprblue}{rgb}{0.21,0.49,0.74}

\twocolumn[{%
\renewcommand\twocolumn[1][]{#1}%
\maketitle
\vspace{-10 mm}
\begin{center}
  \centering
  \includegraphics[width=\linewidth]{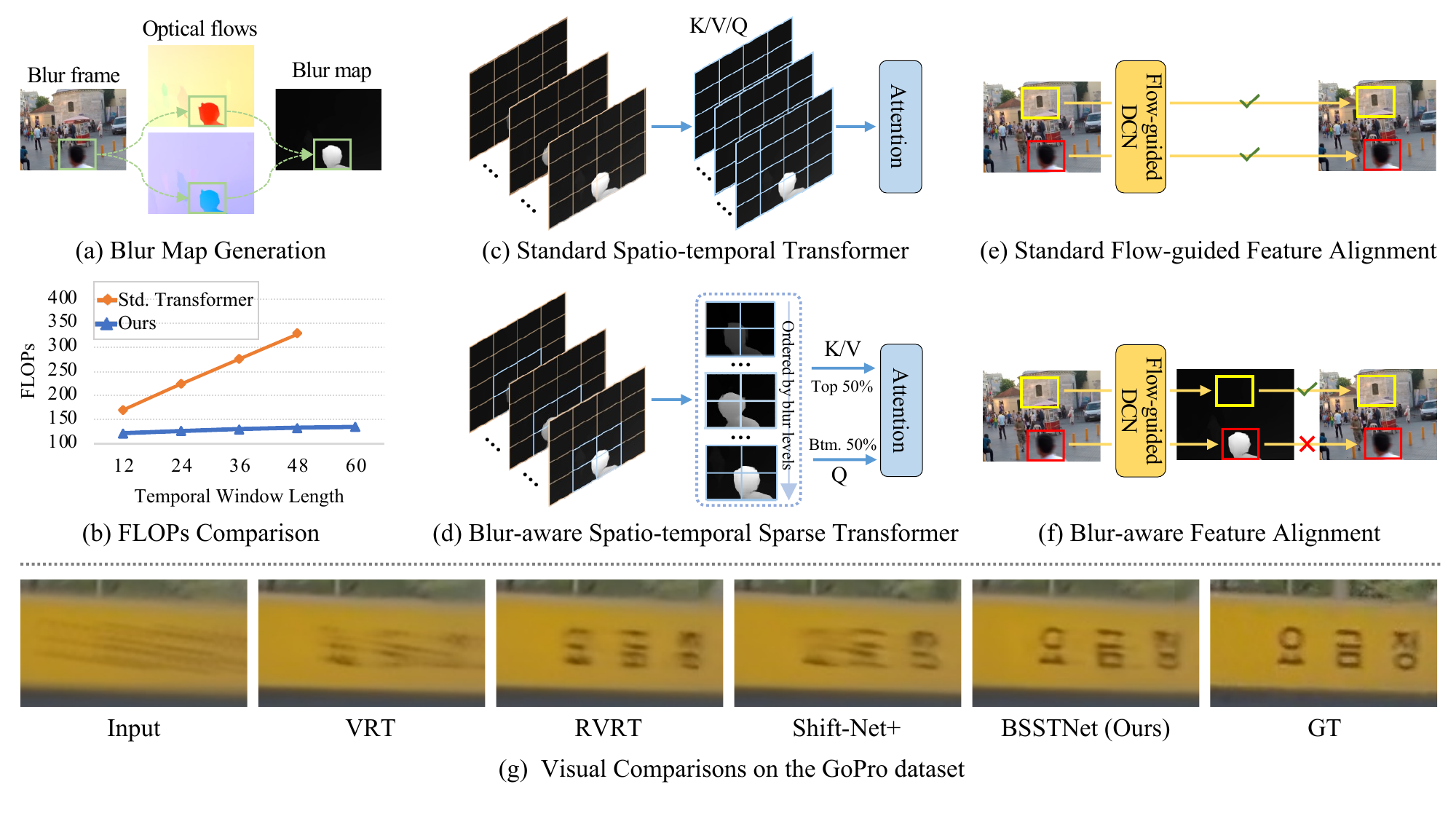}
  \vspace{-5 mm}
  \captionof{figure}{(a) Large motions in optical flows are highlighted in blur maps. (b) Comparison of FLOPs between the standard spatio-temporal transformer and the blur-aware spatio-temporal transformer. (c-d) Summary of the standard spatio-temporal transformer and the blur-aware spatio-temporal transformer.  (e-f) Summary of the standard flow-guided feature alignment and blur-aware feature alignment. (g) In the visual comparisons on the GoPro dataset, the proposed BSSTNet restores the sharpest frame.}
  \label{fig:teaser}
\end{center}%
}]

\renewcommand{\thefootnote}{}
\footnotetext{\textsuperscript{\Letter} Corresponding author: h.yao@hit.edu.cn}

\input{sec/0_abstract}    
\input{sec/1_intro}

\input{sec/2_relatedwork}

\input{sec/3_method}

\input{sec/4_experiments}
\input{sec/5_conclusion}

\noindent \textbf{Acknowledgments}
This research was funded by the National Science and Technology Major Project (2021ZD0110901). 

{
  \small
  \bibliographystyle{ieeenat_fullname}
  \bibliography{main}
}

\end{document}

%% file: sec/0_abstract.tex
\begin{abstract}
Video deblurring relies on leveraging information from other frames in the video sequence to restore the blurred regions in the current frame.
Mainstream approaches employ bidirectional feature propagation, spatio-temporal transformers, or a combination of both to extract information from the video sequence.  
However, limitations in memory and computational resources constraints the temporal window length of the spatio-temporal transformer, preventing the extraction of longer temporal contextual information from the video sequence. 
Additionally, bidirectional feature propagation is highly sensitive to inaccurate optical flow in blurry frames, leading to error accumulation during the propagation process.
To address these issues, we propose \textbf{BSSTNet}, \textbf{B}lur-aware \textbf{S}patio-temporal \textbf{S}parse \textbf{T}ransformer Network.
It introduces the blur map, which converts the originally dense attention into a sparse form, enabling a more extensive utilization of information throughout the entire video sequence. 
Specifically, BSSTNet
(1) uses a longer temporal window in the transformer, leveraging information from more distant frames to restore the blurry pixels in the current frame. 
(2) introduces bidirectional feature propagation guided by blur maps, which reduces error accumulation caused by the blur frame.
The experimental results demonstrate the proposed BSSTNet outperforms the state-of-the-art methods on the GoPro and DVD datasets. 
\end{abstract}

%% file: sec/1_intro.tex
\section{Introduction}
\label{sec:intro}
Video deblurring aims to recover clear videos from blurry inputs, and it finds wide applications in many subsequent vision tasks, including tracking~\cite{DBLP:conf/cvpr/MeiR08,DBLP:conf/cvpr/JinFC05}, video stabilization~\cite{DBLP:journals/pami/MatsushitaOGTS06}, and SLAM~\cite{DBLP:conf/iccv/LeeKL11}. 
Therefore, it is of great interest to develop an effective algorithm to deblur videos for above mentioned high-level vision tasks. 

Video deblurring presents a significant challenge, as it necessitates the extraction of pertinent information from other frames within the video sequence to restore the blurry frame. 
In recent years, there have been noteworthy advancements~\cite{DBLP:journals/corr/abs-2201-12288,DBLP:conf/cvpr/ChanZXL22a,DBLP:conf/cvpr/JiY22,DBLP:conf/icml/LinCHWYZDZTG22,DBLP:conf/eccv/ZhangXY22,DBLP:conf/nips/LiangFXRIGC0TG22} in addressing this challenge. 
Flow-guided bidirectional propagation methods~\cite{DBLP:conf/cvpr/ChanZXL22a,DBLP:conf/cvpr/JiY22,DBLP:conf/icml/LinCHWYZDZTG22,DBLP:conf/eccv/ZhangXY22,DBLP:conf/nips/LiangFXRIGC0TG22} employ flow-guided deformable convolution and flow-guided attention for feature alignment. 
However, inaccurate optical flow in blurry frames causes the introduction of blurry pixels during bidirectional propagation.
VRT and RVRT~\cite{DBLP:journals/corr/abs-2201-12288,DBLP:conf/nips/LiangFXRIGC0TG22} use spatio-temporal self-attention with temporal window to fuse the information from video sequence. 
Due to the high memory demand of self-attention, these approaches frequently feature restricted temporal windows, limiting their ability to incorporate information from distant sections of the video.

Analyzing videos afflicted by motion blur reveals a correspondence between the blurry regions in the video and areas with pixel displacement, where the degree of blurriness is directly associated with the magnitude of pixel displacement. 
Moreover, The blurry regions are typically less frequent in both the temporal and spatial aspects of the blurry videos. 
By leveraging the sparsity of blurry regions, the computation of the spatio-temporal transformer can focus solely on these areas, thereby extending the temporal window to encompass longer video clips.
Moreover, bidirectional feature propagation based on blurry regions enables the minimization of error accumulation. 
As shown in Figure~\ref{fig:teaser}\textcolor{red}{a}, the green box area represents the blurry region in the frame.
Similarly, both the forward and backward optical flows in the same location are also maximized, indicating a correlation between the motion and blurry regions.

By introducing blur maps, we propose \textbf{BSSTNet}, \textbf{B}lur-aware \textbf{S}patio-temporal \textbf{S}parse \textbf{T}ransformer Network. 
Compared to methods based on spatio-temporal transformer, BSSTNet introduces Blur-aware Spatio-temporal Sparse Transformer (BSST) and Blur-aware Bidirectional Feature Propagation (BBFP).
The proposed BSST efficiently utilizes a long temporal window by applying sparsity on input tokens in the spatio-temporal domain based on blur maps.  
This enables the incorporation of distant information in the video sequence while still maintaining computational efficiency.
BBFP introduces guidance from blur maps and checks for flow consistency beforehand.
This aids in minimizing the introduction of blurry pixels during bidirectional propagation, ultimately enhancing the ability to gather information from the adjacent frames.

The contributions are summarized as follows.

\begin{itemize}
\item We propose a non-learnable, parameter-free method for estimating the blur map of video frames. 
The blur map provides crucial prior information on motion-blurry regions in the video, enabling sparsity in the transformer and error correction during bidirectional propagation.
\item  
We propose \textbf{BSSTNet}, comprising two major components: BSST and BBFP. 
BSST incorporates spatio-temporal sparse attention to leverage distant information in the video sequence while still achieving high performance.
BBFP corrects errors in the propagation process and boosts its capability to aggregate information from the video sequence.
\item We quantitatively and qualitatively evaluate BSSTNet on the DVD and GoPro datasets. The experimental results indicate that BSSTNet performs favorably against state-of-the-art methods. 
\end{itemize}

%% file: sec/2_relatedwork.tex
\section{Related Work}
\label{sec:relatedwork}

Many methods in video deblurring have achieved impressive performances. 
The video deblur methods can be categorized into two categories: 

\noindent \textbf{RNN-based Methods.}
On the other hand, some researchers~\citep{DBLP:conf/iccv/KimLSH17,DBLP:conf/iccv/WieschollekHSL17,DBLP:conf/cvpr/NahSL19,DBLP:conf/iccv/ZhouZPZXR19,DBLP:journals/tog/SonLLCL21,DBLP:journals/ijcv/ZhongGZZS23} are focusing on the RNN-base methods. 
STRCNN~\citep{DBLP:conf/iccv/KimLSH17} adopts a recurrent neural network to fuse the concatenation of multi-frame features.
RDN~\citep{DBLP:conf/iccv/WieschollekHSL17} develops a recurrent network to recurrently use features from the previous frame at multiple scales.
IFRNN~\citep{DBLP:conf/cvpr/NahSL19} adopts an iterative recurrent neural network (RNN) for video deblurring.
STFAN~\citep{DBLP:conf/iccv/ZhouZPZXR19} uses dynamic filters to align consecutive frames.
PVDNet~\citep{DBLP:journals/tog/SonLLCL21} contains a pre-trained blur-invariant flow estimator and a pixel volume module.
To aggregate video frame information, ESTRNN~\citep{DBLP:journals/ijcv/ZhongGZZS23} employs a GSA module in the recurrent network.
Recently, the BiRNN-based method~\citep{DBLP:conf/cvpr/ChanZXL22a,DBLP:conf/aaai/ZhuDPLHFW22,DBLP:conf/cvpr/JiY22,DBLP:conf/icml/LinCHWYZDZTG22,DBLP:conf/eccv/ZhangXY22} has achieved impressive deblur results through aggressive bidirectional propagation.
BasicVSR$++$\citep{DBLP:conf/cvpr/ChanZXL22a} adopts aggressive bidirectional propagation.
Based on BasicVSR$++$, RNN-MBP\citep{DBLP:conf/aaai/ZhuDPLHFW22} introduces the multi-scale bidirectional recurrent neural network for video deblurring. 
STDANet~\cite{DBLP:conf/eccv/ZhangXY22} and FGST~\cite{DBLP:conf/icml/LinCHWYZDZTG22} employ the flow-guided attention to align and fuse the information of adjacent frames. 
However, due to error accumulation, these methods do not effectively fuse the information from long-term frames.
Ji and Yao~\citep{DBLP:conf/cvpr/JiY22} develop a Memory-Based network, which contains a multi-scale bidirectional recurrent neural network and a memory branch.
However, the memory branch introduces a large search space of global attention and ineffective alignment.  

\begin{figure*}[!t]
  \includegraphics[width=\linewidth]{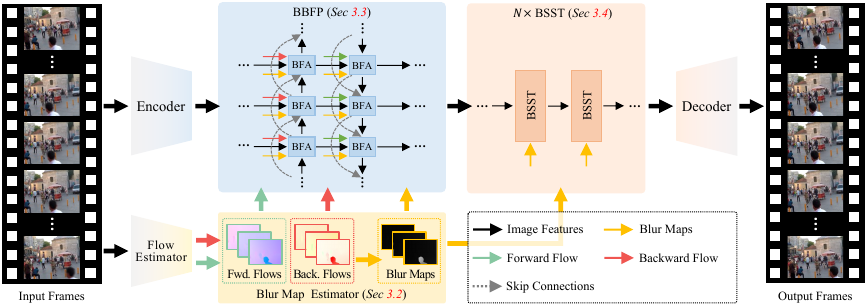}
  \caption{\textbf{Overview of the proposed BSSTNet.} BSSTNet consists of three major components: Blur Map Estimation, Blur-aware Bidirectional Feature Propagation (BBFP), and Blur-aware Spatio-temporal Sparse Transformer (BSST).}
  \label{fig:overview}
\end{figure*}

\noindent \textbf{Transformer-based Methods.} 
The Spatio-temporal transformer is widely used in video deblurring~\cite{DBLP:journals/corr/abs-2201-12288,DBLP:conf/nips/LiangFXRIGC0TG22}. 
VRT~\cite{DBLP:journals/corr/abs-2201-12288} utilizes spatio-temporal self-attention mechanism to integrate information across video frames. Due to the computational complexity of self-attention, VRT employs a 2-frame temporal window size and utilizes a shifted window mechanism for cross-window connections. 
However, the indirect connection approach with a small window size fails to fully exploit long-range information within the video sequence. 
RVRT~\cite{DBLP:conf/nips/LiangFXRIGC0TG22}  divides the video sequence into 2-frame clips, employing small-window spatio-temporal self-attention within each clip and Flow-guided biderectional propagation and alignment between clips. 
However, due to the small window constraint of spatio-temporal self-attention and the error accumulation caused by the optical flow of blurred frames in Flow-guided biderectional propagation, RVRT still falls short of fully utilizing the information from the entire video sequence.

%% file: sec/3_method.tex
\section{Our Approach} 
\label{sec:method}
\subsection{Overview} 

%
As shown in Figure~\ref{fig:overview}, the BSSTNet contains three key components:
Blur Map Estimation, Blur-aware Bidirectional Feature Propagation (BBFP), and Blur-aware Spatio-temporal Sparse Transformer (BSST). 
First, the forward and backward optical flows, denoted as $\{\mathbf{O}_{t+1\rightarrow t}\}_{t=1}^{T-1}$ and $\{\mathbf{O}_{t\rightarrow t+1}\}_{t=1}^{T-1}$, are estimated from the downsampled video sequence $\hat{\mathbf{X}}=\{\hat{\mathbf{X}}_{t}\}_{t=1}^T$. 
Then, Blur Map Estimation generates the blur maps  $\mathcal{B} = \{\mathbf{B}_t\}_{t=1}^T$ for each frame are generated based on $\{\mathbf{O}_{t+1\rightarrow t}\}_{t=1}^{T-1}$ and $\{\mathbf{O}_{t\rightarrow t+1}\}_{t=1}^{T-1}$. 
Next, BBFP produces the aggregated features $\hat{\mathbf{F}}$ using Blur-aware Feature Alignment (BFA).  
After that, BSST generates the refined features $\overline{\mathbf{F}}$ from $\hat{\mathbf{F}}$ with the \textbf{B}lur-aware \textbf{S}parse \textbf{S}patio-temporal \textbf{A}ttention (BSSA) layers. 
Finally, the decoder reconstructs the sharp video sequence $\mathcal{R} = \{\mathbf{R}_{t}\}_{t=1}^T$. 

\subsection{Blur Map Estimation}
\label{sec:Blur_map_es_method}

Given the optical flows $\{\mathbf{O}_{t+1\rightarrow t}\}_{t=1}^{T-1}$ and $\{\mathbf{O}_{t\rightarrow t+1}\}_{t=1}^{T-1}$, 
the unnormalized blur maps $\hat{\mathcal{B}} = \{\hat{\mathbf{B}}_t\}_{t = 1}^T$ can be obtained as follows
\begin{align}
\hat{\mathbf{B}}_t &= \sum\limits_{i=1}^{2} ( ( \mathbf{O}_{t\rightarrow t+1} )^2_i + ( \mathbf{O}_{t\rightarrow t-1} )^2_i) 
\end{align}
Specially, we define $\mathbf{O}_{1 \rightarrow 0} = \mathbf{0}$ and $\mathbf{O}_{T \rightarrow T+1} = \mathbf{0}$.
The blur map $\mathbf{B}$ and sharp map $\mathbf{A}$ can be generated as follows
\begin{align}
\mathbf{B}_t &= \frac{\hat{\mathbf{B}_t} - \min(\hat{\mathcal{B}})} 
{\max(\hat{\mathcal{B}}) - \min({\hat{\mathcal{B}})}} 
\nonumber \\
\mathbf{A}_t &= 1 - \mathbf{B}_t
\label{eq:blur_map}
\end{align}
where $i$ and $t$ index the channel of optical flows and the time steps, respectively.

\subsection{Blur-aware Bidirectional Feature Propagation}
\label{sec:BBFP}

Within BBFP, bidirectional feature propagation propagates the aggregated features $\hat{\mathbf{F}}$ in both the forward and backward directions, incorporating Blur-aware Feature Alignment (BFA).
BFA is designed to align features from neighboring frames to reconstruct the current frame. 
As shown in Figure~\ref{fig:teaser}\textcolor{red}{e} and Figure~\ref{fig:teaser}\textcolor{red}{f}, the standard flow-guided feature alignment aligns all pixels in the neighboring frames, whereas BFA selectively integrates information from sharp pixels guided by blur maps.
This prevents the propagation of blurry regions from the features of neighboring frames during bidirectional feature propagation.

\noindent \textbf{Bidirectional Feature Propagation.}
Assuming the current time step is the $t$-th step, and the corresponding propagation branch is the $j$-th branch, 
the generation of the current time step aggregated feature $\hat{\mathbf{F}}_{t}^j$ can be obtained as
\begin{align}
  \hat{\mathbf{F}}_{t}^j = \textrm{BFA}(
    &\hat{\mathbf{F}}_t^{j-1},\hat{\mathbf{F}}_{t-1}^{j},\hat{\mathbf{F}}_{t-2}^{j}, \nonumber \\
    &\textrm{W}(\hat{\mathbf{F}}_{t-1}^{j},\mathbf{O}_{ t \rightarrow t-1}), \textrm{W}(\hat{\mathbf{F}}_{t-2}^{j}, \mathbf{O}_{ t \rightarrow t-2}), \nonumber \\
    &\mathbf{O}_{ t \rightarrow t-1},\mathbf{O}_{t \rightarrow t-2},\mathbf{A}_{t-1},\mathbf{A}_{t-2})
\end{align}
where $\textrm{BFA}$ and $\textrm{W}$ denote the ``BFA'' and ``Backward Warp'' operations, resepectively.
$\hat{\mathbf{F}}_t^{j-1}$ represents the feature aggregated from the $t$-th time step in the $(j-1)$-th branch.
$\hat{\mathbf{F}}_{t-1}^{j}$ and $\hat{\mathbf{F}}_{t-2}^{j}$ are the features generated from the previous and the second previous time step.
The aforementioned process progresses forward through the time steps until it reaches $t = T$.
The backward propagation process mirrors the forward propagation process.

\begin{figure}[!t]
  \includegraphics[width=\linewidth]{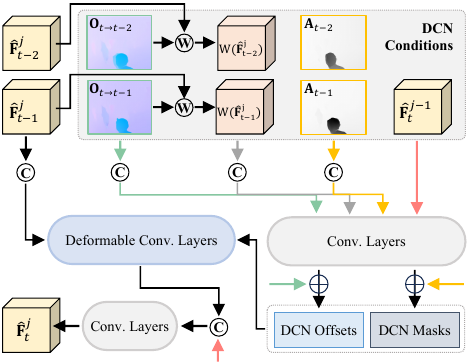}
  \caption{\textbf{The details of BFA.} Note that \raisebox{.5pt}{\textcircled{\raisebox{-.9pt} {W}}}, \raisebox{.5pt}{\textcircled{\raisebox{-.9pt} {C}}}, and $\bigoplus$ denotes the ``Warp'', ``Concatenation'', and ``Element-wise Add'' operations, respectively.}
  \label{fig:BFA}
  \vspace{-2 mm}
\end{figure}

\noindent {\bf Blur-aware Feature Alignment.}
Different from the standard flow-guided feature alignment~\cite{DBLP:conf/cvpr/ChanZXL22a} that aligns all pixels in neighboring frames, BFA introduces sharp maps to prevent the introduction of blurry pixels in the neighboring frames.
As illustrated in Figure~\ref{fig:BFA}, along with 
features $\hat{\mathbf{F}}_{t - 1}^j$ and $\hat{\mathbf{F}}_{t - 2}^j$ from previous time steps, 
the corresponding optical flows $\mathbf{O}_{t \rightarrow t-1}$ and $\mathbf{O}_{t \rightarrow t-2}$, 
and the warped features ${\rm W}(\hat{\mathbf{F}}_{t - 1}^j)$ and ${\rm W}(\hat{\mathbf{F}}_{t - 2}^j)$,
sharp maps $\mathbf{A}_{t - 1}$ and $\mathbf{A}_{t - 2}$ are additionally introduced.
These sharp maps serve as additional conditions to generate the offsets and masks of the deformable convolution layers~\cite{DBLP:conf/iccv/DaiQXLZHW17}.
Moreover, the sharp map acts as a base mask for DCN by being added to the DCN mask. 
This ensures that only sharp regions of features are propagated.

\subsection{Blur-aware Spatio-temporal Sparse Transformer}
\label{sec:BSST}

\begin{figure}[!t]
  \includegraphics[width=\linewidth]{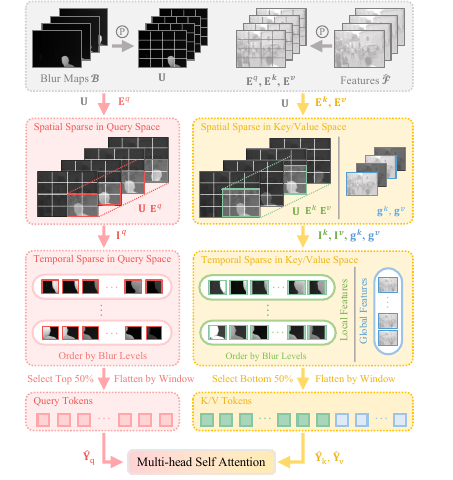}
  \caption{\textbf{The details of BSST.} Note that \raisebox{.5pt}{\textcircled{\raisebox{-.9pt} {P}}} denotes the ``Window Partition'' operation. ``Flatten by window'' indicates that query tokens are flattened for each query window, and K/V tokens are generated in a similar manner. Multi-head Self Attention is also computed on the query and K/V tokens generated for each window.}
  \label{fig:bsst}
  \vspace{-4 mm}
\end{figure}

The spatio-temporal attention is commonly employed in video deblurring and demonstrates remarkable performance, as shown in Figure~\ref{fig:teaser}\textcolor{red}{c}.
However, the standard spatio-temporal attention method often restricts its temporal window size due to computational complexity, thereby constraining its capability to capture information from distant parts of the video sequence.
To overcome this limitation, we introduce the Blur-aware Spatio-temporal Sparse Transformer (BSST).
As illustrated in Figure~\ref{fig:teaser}\textcolor{red}{d}, BSST filters out unnecessary and redundant tokens in the spatio and temporal domain according to blur maps $\mathbf{B}$.
As shown in Figure~\ref{fig:teaser}\textcolor{red}{b}, allowing BSST to include a larger temporal window while maintaining computational efficiency. 
The detailed implementation of BSST is illustrated in Figure~\ref{fig:bsst}.

Given the aggregated features $\hat{\mathcal{F}} = \{ \hat{\mathbf{F}}_{t} \in \mathbb{R}^{H/4 \times W/4 \times C} \}_{t=1}^{T}$ from the last branch of BBFP, 
 
we employ a soft split operation~\cite{DBLP:conf/iccv/0019DHSLS0D021} to divide each aggregated feature into overlapping patches of size $p \times p$ with a stride of $s$. 
The split features are then concatenated, generating the patch embeddings $\mathbf{z} \in \mathbb{R}^{T \times M \times N \times p^{2}C}$. 
Next, the blur map $\mathbf{B}$ are downsampled by average pooling with a kernel size of $p \times p$ and a stride of $s$, resulting in $\mathbf{B}^{\downarrow} \in \mathbb{R}^{T \times M \times N \times p^{2}C}$. 
For simplicity, $p^{2}C$ is denoted as $C_z$. 
After that, $\mathbf{z}$ is fed to three separate linear layer transformations, resulting in $\Tilde{\mathbf{z}}^q \in \mathbb{R}^{T \times M \times N \times C_z}$, $\Tilde{\mathbf{z}}^k \in \mathbb{R}^{T \times M \times N \times C_z}$, and $\Tilde{\mathbf{z}}^v \in \mathbb{R}^{T \times M \times N \times C_z}$, 
where $M$, $N$, and $C_z$ respectively denote the number of patches in the height and width domains, and the number of channels.
Subsequently, $\Tilde{\mathbf{z}}^q, \Tilde{\mathbf{z}}^k, \Tilde{\mathbf{z}}^v$ are partitioned into $m\times n$ non-overlapping windows, generating partitioned features $ \mathbf{E}^q, \mathbf{G}^k, \mathbf{G}^v  \in \mathbb{R}^{T \times m\times n \times h\times w\times C_z}$, 
where $m\times n$ and $h\times w$ are the number and size of the windows, respectively. 
Utilizing the embedding $\mathbf{z}$ and incorporating depth-wise convolution, the generation of pooled global tokens $\mathbf{g}^k$ and $\mathbf{g}^v$ takes place as follows
\begin{align}
  \mathbf{g}^k &= { \rm l_k}(\textrm{DC}(z)) \nonumber \\
  \mathbf{g}^v &= {\rm l_v}(\textrm{DC}(z)) 
  \label{eq:pooling_tokens}
\end{align}
where $\textrm{DC}$ represents depth-wise convolution, and $\mathbf{g}^k, \mathbf{g}^v \in \mathbb{R}^{T \times h_p \times w_p \times C_z}$. 
Following that, we repeat and concatenate $\mathbf{g}^k$ with $\mathbf{G}^k$ and $\mathbf{g}^v$ with $\mathbf{G}^v$, resulting in $\mathbf{E}^k, \mathbf{E}^v \in \mathbb{R}^{T \times m \times n \times (h + h_p) \times (w + w_p) \times C_z}$. 
Note that for the key/value windows, we enlarge its window size to enhance the receptive field of key/value~\cite{DBLP:conf/iccv/0019DHSLS0D021,DBLP:conf/iccv/ZhouLCL23}. 
For simplicity, we ignore it in the following discussion.

\noindent {\bf Spatial Sparse in Query/Key/Value Spaces.}
We observe that the blurry regions are typically less frequent in both the temporal and spatial aspects of the blurred videos. 
Motivated by this observation, we only choose the tokens of blurry windows in $\mathbf{E}^q$ and tokens of sharp windows in $\mathbf{E}^k,\mathbf{E}^v$ to participate in the computation of spatio-temporal attention.
This ensures that the spatio-temporal attention mechanism focuses solely on restoring the blurred regions by the utilization of sharp regions in video sequences. 
First, the blur maps of windows $\mathbf{U} \in \mathbb{R}^{T \times m \times n}$ are generated by downsampling $\mathbf{B}^{\downarrow} \in \mathbb{R}^{T \times M \times N}$ using max pooling.
Next , the spatial sparse mask of windows is obtained as follows
\begin{align}
\mathbf{Q}_{t,i,j} &= 
\begin{cases} 
1, & \text{if } \mathbf{U}_{t,i,j} \geq \theta,  \nonumber \\
&\forall t \in [1, T], i \in [1, m], j \in [1, n] \\
0, & \text{otherwise} 
\end{cases} \\
\mathbf{S} &= {\rm Clip}~\Big(\sum \nolimits_{t=1}^{T} \mathbf{Q}_t, ~1\Big)
\label{eq:blur_mask}
\end{align}
where $\theta, \textrm{Clip}, \mathbf{S} \in \mathbb{R}^{m \times n}$ are the threshold for considering related windows as blurry windows, a clipping function that set $\mathbf{S}$ to 1 if $\sum \nolimits_{t=1}^{T} {\mathbf{Q}}_t > 0$, and the spatial sparse mask for $\mathbf{E}^q$, $\mathbf{E}^k$ and $\mathbf{E}^v$, respectively. 
Then, the spatial sparse embedding features $ \mathbf{I}^q$, $\mathbf{I}^k$, and $\mathbf{I}^v$ are generated using the following equations
\begin{align}
\mathbf{I}^q &= {\rm Concat} ( \{ \mathbf{E}^q_{t,i,j} \mid \mathbf{S}_{i,j} = 1, i \in [1, m], j \in [1, n] \}_{t=1}^T )
\nonumber \\
\mathbf{I}^k &= {\rm Concat} ( \{ \mathbf{E}^k_{t,i,j} \mid \mathbf{S}_{i,j} = 1, i \in [1, m], j \in [1, n] \}_{t=1}^T )
\nonumber \\
\mathbf{I}^v &= {\rm Concat} ( \{ \mathbf{E}^v_{t,i,j} \mid \mathbf{S}_{i,j} = 1, i \in [1, m], j \in [1, n] \}_{t=1}^T ) 
\label{eq:spatialsparse_qkv}
\end{align}
where ${\rm Concat}$ denotes the ``Concatenation'' operation. 
If $S_{i, j} = 0$, it indicates that the window's position indexed by $(i, j)$ in the video sequence does not encompass blurry tokens. 
This allows us to exclude the tokens within those windows from the spatio-temporal attention mechanism.  
$\mathbf{I}^q \in \mathbb{R}^{T \times m_s n_s \times h  w \times C_z}$, while both $\mathbf{I}^k$ and $\mathbf{I}^v$ share the size of $\mathbb{R}^{T \times m_s  n_s \times (h + h_p) (w + w_p)\times C_z}$, where $m_s$ and $n_s$ representing the number of selected windows in $m$ and $n$ domains, respectively. 

\noindent {\bf Temporal Sparse in Query Space.}
Along the temporal domain, we choose the windows of the blurry region for query space, ensuring that the spatio-temporal attention mechanism is dedicated to restoring only the blurry regions of the video sequence. 
Given the spatial sparse embedding features $\mathbf{I}^q$, the spatio-temporal sparse embedding $\mathbf{y}^q$ is generated as follows
\begin{align}
  \mathcal{H}^q &= \{ \mathbf{I}^q_{t,i,j} \mid \mathbf{U}_{t,i,j} \geq {\rm Top}(K^q, \mathbf{U}_{i,j}), \nonumber \\ 
   & i \in [1, m_s], j \in [1, n_s] \}_{t=1}^T \nonumber \\ 
  \mathbf{Y}^q &= {\rm Concat}(\mathcal{H}^q)
  \label{eq:temporalsparse_q}
\end{align}
where $\mathbf{Y}^q \in \mathbb{R}^{K_q \times m_s n_s \times h w \times C_z}$.
$\textrm{Top}(K^q, \cdot)$ represents the operation of finding the $K_q$-th largest element in a vector. 
For each window located at position $(i, j)$ in $\mathbf{I}^q$, within the temporal domain, we selectively chose the top $K_q$ windows with the highest blur levels for deblurring.

\begin{table*}[!t]
\caption{\textbf{Quantitative comparisons on the GoPro dataset.} The best results are highlighted in bold.}
\vspace{-2 mm}
\resizebox{\linewidth}{!} {
	\begin{tabular}{lccccccccc}
		\toprule
		Method
		& STFAN~\cite{DBLP:conf/iccv/ZhouZPZXR19}
		& STDAN~\cite{DBLP:conf/eccv/ZhangXY22}
            & RNN-MBP~\cite{DBLP:conf/aaai/ZhuDPLHFW22}
		& NAFNet~\cite{DBLP:conf/eccv/ChenCZS22}
		
		& VRT~\cite{DBLP:journals/corr/abs-2201-12288}
		& RVRT~\cite{DBLP:conf/nips/LiangFXRIGC0TG22}
		& Shift-Net+~\cite{DBLP:conf/cvpr/LiSZCSWQL23}
		& BSSTNet \\
		\midrule
		PSNR     & 28.69      & 32.62     & 33.32  & 33.69    & 34.81    & 34.92  & 35.88   & \bf{35.98} \\
		SSIM      & 0.8610      & 0.9375      & 0.9627 & 0.9670   & 0.9724    & 0.9738   & 0.9790   & \bf{0.9792}  \\
		\bottomrule
	\end{tabular}
}
\label{tab:psnr_GoPro}
\end{table*}

\begin{table*}[t]
\caption{\textbf{Quantitative comparisons on the DVD dataset.} The best results are highlighted in bold.}
\vspace{-2 mm}
\resizebox{\linewidth}{!} {
	\begin{tabular}{lccccccccc}
		\toprule
		Method
		& STFAN~\cite{DBLP:conf/iccv/ZhouZPZXR19}
            & ARVo~\cite{DBLP:conf/cvpr/LiXZ0ZRSL21}
		& RNN-MBP~\cite{DBLP:conf/aaai/ZhuDPLHFW22}
            & STDAN~\cite{DBLP:conf/eccv/ZhangXY22}
		& VRT~\cite{DBLP:journals/corr/abs-2201-12288}
		& RVRT~\cite{DBLP:conf/nips/LiangFXRIGC0TG22}
		& Shift-Net+~\cite{DBLP:conf/cvpr/LiSZCSWQL23}
		& BSSTNet \\
		\midrule
		PSNR     & 31.24      & 32.80       & 32.49    & 33.05     & 34.27    & 34.30  & 34.69   & \bf{34.95} \\
		SSIM      & 0.9340      & 0.9352      & 0.9568      & 0.9374    & 0.9651    & 0.9655   & 0.9690   & \bf{0.9703}  \\
		\bottomrule
	\end{tabular}
}
\label{tab:psnr_dvd}
\end{table*} 

\noindent {\bf Temporal Sparse in Key/Value Spaces.}
In contrast to the query space, we select the sharp regions in $\mathbf{I}_k, \mathbf{I}_v$ for key/value spaces.  
Due to the high similarity in textures between adjacent frames, we alternately choose temporal frames with a stride of 2 in each BSST. 
In BSSTNet, consisting of multiple BSSTs, odd-numbered BSSTs select frames with odd numbers, while even-numbered BSSTs choose frames with even numbers, resulting in a 50\% reduction in the size of the key/value space.
Given the spatial sparse embedding features $\mathbf{I}^k$ and $\mathbf{I}^v$, the spatio-temporal sparse embedding features $\mathbf{y}^k$ and $\mathbf{y}^v$ are generated as follows
\begin{align}
\mathcal{H}^k &= \{ \mathbf{I}^k_{t,i,j} \mid \mathbf{U}_{t,i,j} \geq {\rm Top}(K_{kv}, 1 - \mathbf{U}_{i,j}), \nonumber \\ 
 &t \mod 2 = 0, i \in [1, m_s], j \in [1, n_s] \}_{t=1}^T \nonumber \\
 \mathbf{y}^k &= {\rm Concat}(\mathcal{H}^k) \nonumber \\
 \mathcal{H}^v &= \{ \mathbf{I}^v_{t,i,j} \mid \mathbf{U}_{t,i,j} \geq {\rm Top}(K_{kv}, 1- \mathbf{U}_{i,j}), \nonumber \\ 
 &t \mod 2 = 0, i \in [1, m_s], j \in [1, n_s] \}_{t=1}^T  \nonumber \\
  \mathbf{y}^v &= {\rm Concat}(\mathcal{H}^v) 
\label{eq:temporalsparse_kv}
\end{align}
where $\mathbf{y}^k, \mathbf{y}^v \in \mathbb{R}^{K_{kv} \times m_s n_s \times  (h + h_p)(w + w_p) \times C_z}$. 

\noindent {\bf Spatio-temporal Sparse Attention.}
The spatio-temporal sparse query embedding $\mathbf{y}^q$ is reshaped into $\hat{\mathbf{Y}}_q \in \mathbb{R}^{m_s n_s \times K_qhw \times C_z}$. 
Similarly, The spatio-temporal sparse key/value embedding $\mathbf{y}^k$ and $\mathbf{y}^v$ are each reshaped into $\hat{\mathbf{Y}}_k \in \mathbb{R}^{m_s  n_s \times K_{kv}(h + h_p )(w + w_p ) \times C_z}$ and $\hat{\mathbf{Y}}_v \in \mathbb{R}^{m_s  n_s \times K_{kv}(h + h_p )(w + w_p ) \times C_z}$, respectively. 
For each window in $m_sn_s$, the self-attention is calculated as follows:
\begin{align}
\text{Attention}(\hat{\mathbf{Y}}_q,\hat{\mathbf{Y}}_k,\hat{\mathbf{Y}}_v) = {\rm Softmax}\left(\frac{\hat{\mathbf{Y}}_q\hat{\mathbf{Y}}_k^T}{\sqrt{C_z}}\right) \hat{\mathbf{Y}}_v
\label{eq:attn}
\end{align}
In BSST, the multi-head self-attention is introduced to obtain the output embedding $\mathbf{z}_s \in 
\mathbb{R}^{m_s \times n_s \times K_qhw \times C_z}$.
\begin{align}
\mathbf{z}_s = {\rm MSA}(\hat{\mathbf{Y}}_q,\hat{\mathbf{Y}}_k,\hat{\mathbf{Y}}_v) 
\label{eq:msa}
\end{align}
where ${\rm MSA}$ is the ``Multi head Self-Attention'' function.
After applying our sparse strategy to eliminate unnecessary and redundant windows, we use self-attention following Eq.~\ref{eq:attn} on the remaining windows to extract fused features. 
Specially, standard window spatio-temporal attention is applied to unselected (less blurry) windows, allowing features to be restored to their original size. 
Subsequently, these features are gathered through a soft composition operation~\cite{DBLP:conf/iccv/0019DHSLS0D021} to serve as the input for the next BSST.
The output of the final BSST is denoted as $\overline{\mathbf{F}}$.

%% file: sec/4_experiments.tex
\section{Experiments}

\subsection{Datasets}

\noindent \textbf{DVD}.
The DVD dataset~\cite{DBLP:conf/cvpr/SuDWSHW17} comprises 71 videos, consisting of 6,708 blurry-sharp pairs. These are divided into 61 training videos, amounting to 5,708 pairs, and 10 testing videos with 1,000 pairs. 

\noindent \textbf{GoPro}.
The GoPro dataset~\cite{DBLP:conf/cvpr/NahKL17} consists of 3,214 pairs of blurry and sharp images at a resolution of 1280$\times$720. Specifically, 2,103 pairs are allocated for training, while 1,111 pairs are designated for testing. 

\subsection{Implementation Details}

\noindent \textbf{Training Details}
The network is implemented with PyTorch~\cite{DBLP:conf/nips/PaszkeGMLBCKLGA19}~\footnote{The source code is available at \url{https://github.com/huicongzhang/BSSTNet}}.
The training is conducted with a batch size of 8 on 8 NVIDIA A100 GPUs, and the initial learning rate is set to $4 \times 10^{-4}$. 
The network is optimized with L1 loss using Adam optimizer~\cite{DBLP:journals/corr/KingmaB14}, where $\beta_1 = 0.9$ and $\beta_2 = 0.999$. 
The flow estimator in BSSTNet uses pre-trained weights from the official RAFT~\cite{DBLP:conf/eccv/TeedD20} release and remains fixed during training. 
During testing, $T$, $K_q$, and $K_{kv}$ are set to $48, 24,$ and $24$, respectively. 
During training, they are $24, 12,$ and $12$, respectively. 
In the training phase, input images are randomly cropped into patches with resolutions of $256 \times 256$, along with the application of random flipping and rotation. 

\noindent \textbf{Hyperparameters}
To strike a better balance between video deblurring quality and computational efficiency, the value of $\theta$ is set to $0.3$. 
The patch size $p$ and stride $z$ are set to $4$ and $2$, respectively.

\begin{table}[t]
  \caption{\textbf{The comparison of FLOPs and runtime on the DVD dataset.} The top two results are marked in bold and underlined. Note that FLOPs and runtime are computed for a single frame with a resolution of $ 256 \times 256$.}
  \resizebox{\linewidth}{!} {
	\begin{tabular}{lcccc}
		\toprule
		Method
		& RVRT~\cite{DBLP:conf/nips/LiangFXRIGC0TG22}
		& Shift-Net+~\cite{DBLP:conf/cvpr/LiSZCSWQL23}
		& BSSTNet \\
		\midrule
		PSNR       & 34.30  & 34.69    & \bf{34.95} \\
		SSIM          & 0.9655   & 0.9690    & \bf{0.9703}  \\
         \midrule
         GFLOPs & \bf{88.8} & 146  & \underline{133} \\
         Runtime (ms) & \bf{23} & 45  & \underline{28} \\
		\bottomrule
	\end{tabular}
  }
  \label{tab:flops_runtime}
  \vspace{-4 mm}
\end{table} 

\begin{figure*}[!t]
  \begin{subfigure}[t]{\textwidth}
    \includegraphics{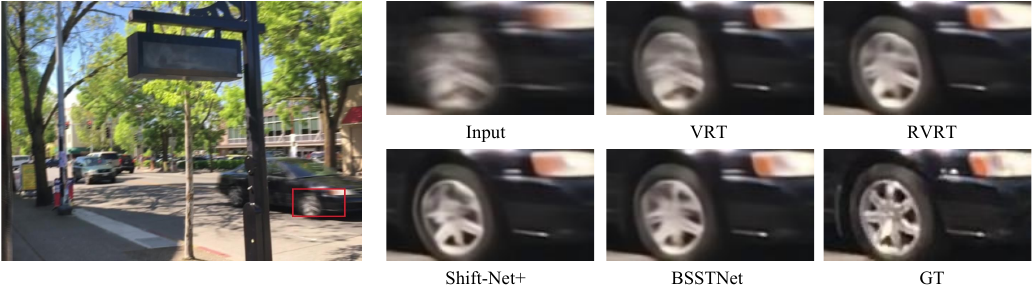}
    \vspace{-6 mm}
    \caption{Qualitative comparison on the DVD dataset}
    \vspace{2 mm}
    \label{fig:quali_dvd}
  \end{subfigure}
  \begin{subfigure}[t]{\textwidth}
    \includegraphics{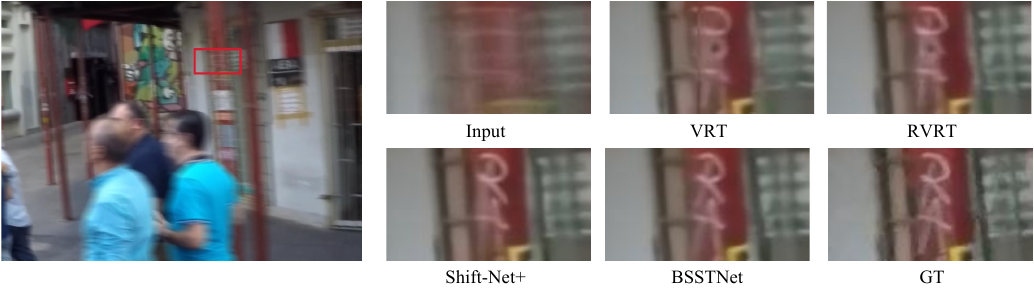}
    \vspace{-6 mm}
    \caption{Qualitative comparison on the GoPro dataset}
    \vspace{4 mm}
    \label{fig:quali_gopro}
  \end{subfigure}
  \vspace{-6 mm}
  \caption{\textbf{Qualitative comparison on the GoPro and DVD datasets.} Note that ``GT'' stands for ``Ground Truth''. The proposed BSSTNet produces images with enhanced sharpness and more detailed visuals compared to competing methods.}
  \vspace{-2 mm}
\end{figure*}

\subsection{Main Results}

\noindent \textbf{DVD.} 
The quantitative results on the DVD dataset are shown in Table~\ref{tab:psnr_dvd}. 
The proposed method demonstrates superior performance in terms of both PSNR and SSIM compared to existing state-of-the-art methods. 
Specifically, in comparison to the best-performing state-of-the-art method, Shift-Net+, the proposed BSSTNet achieves an improvement of \textbf{0.26 dB}  in PSNR and \textbf{0.0013} in SSIM.
Examples from the DVD dataset are presented in Figure~\ref{fig:quali_dvd}, demonstrating that the proposed method generates images with increased sharpness and richer visual details. 
This highlights the robustness of the method in eliminating large blur in dynamic scenes. 

\noindent \textbf{GoPro.}
In Table~\ref{tab:psnr_GoPro}, the proposed BSSTNet shows favorable performance in terms of both PSNR and SSIM when compared to state-of-the-art methods on the GoPro dataset. BSSTNet achieves higher PSNR and SSIM values compared to Shift-Net+.
The visual results in Figure~\ref{fig:quali_gopro} further illustrate that the proposed method restores finer image details and structures.

\noindent \textbf{FLOPs and Runtime.} 
We conducted a comparison of the computational complexity (FLOPs) and runtime between our method, RVRT, and Shift-Net+, as presented in Table~\ref{tab:flops_runtime}. 
In contrast to the state-of-the-art Shift-Net+, our approach demonstrates a \textbf{13 GFLOPs} reduction in FLOPs and achieves a speedup of \textbf{1.6} times.  

\begin{table*}[!t]
  \centering
  \caption{\textbf{Comparison of different temporal lengths in terms of PSNR, SSIM, Runtime, Memory, and GFLOPs between the Standard Spatio-temporal Transformer (SST) and BSST.}. The results are evaluated on the DVD dataset. Note that ``TL.'' and ``Mem.'' denote ``Temporal Length'' and the used memory on GPU, respectively. SST runs out of memory for a temporal length of 60. }
  \vspace{-2 mm}
  \begin{tabularx}{\linewidth}{Y|YYccc|YYccc}  
    \toprule
      \multirow{2}{*}{TL.} & 
      \multicolumn{5}{c|}{SST} & 
      \multicolumn{5}{c}{BSST} \\
      \cline{2-6} \cline{7-11}
          & PSNR  & SSIM   & Time (ms) & Mem. (GB) & GFLOPs 
          & PSNR  & SSIM   & Time (ms) & Mem. (GB) & GFLOPs \\
      \midrule
      12  & 34.59 & 0.9684 & 470       & 6.35      & 171 
          & 34.52 & 0.9681 & 336       & 2.40      & 122 \\
      24  & 34.83 & 0.9696 & 925       & 13.10     & 251 
          & 34.74 & 0.9692 & 684       & 4.79      & 127 \\
      36  & 34.92 & 0.9702 & 1332      & 20.40     & 277 
          & 34.85 & 0.9697 & 1026      & 7.30      & 130 \\
      48  & 34.97 & 0.9704 & 1776      & 28.27     & 329 
          & 34.95 & 0.9703 & 1368      & 9.97      &  133\\
      60  & -     & -      & -         & -         & - 
          & 35.01 & 0.9706 & 1712      & 12.78     & 137 \\
      \bottomrule
    \end{tabularx}
  \label{tab:temporal_length}
  \vspace{-2 mm}
\end{table*}

\begin{figure}[!t]
  \centering
  \includegraphics[width=\linewidth]{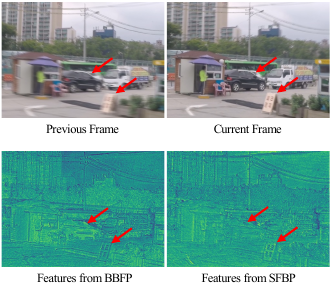}
  \caption{\textbf{Comparison of feature alignment between BBFP and Standard Flow-guided Bidirectional Propagation (SFBP).} Compared to SFBP, BBFP prevents the propagation of blurry regions from the features of neighboring frames during propagation.}
  \label{fig:ab_bbfp}
  \vspace{-2 mm}
\end{figure}

\begin{table}[!t]
  \centering
  \caption{\textbf{Effectiveness of BBFP and BSST.} The best results are highlighted in bold. The results are evaluated on the DVD dataset.}
  \vspace{-2 mm}
  \begin{tabularx}{\linewidth}{Y|YYYY}
    \toprule
    Exp. & (a) & (b) & (c) & (d) \\
    \midrule
    BBFP &  &            & \checkmark & \checkmark \\
    BSST &  & \checkmark &            & \checkmark \\
    \midrule
    PSNR          & 33.78       & 34.74       & 34.10         & \bf{34.95}      \\
    SSIM           & 0.9645      & 0.9692       & 0.9661          & \bf{0.9703}    \\
    \bottomrule
  \end{tabularx}
  \label{tab:structure}
  \vspace{-2 mm}
\end{table} 

\begin{table}[!t]
  \centering
  \caption{\textbf{Comparison between BFA and Standard Flow-guided Feature Alignment (SFFA).} The best results are highlighted in bold. The results are evaluated on the DVD dataset.}
  \vspace{-2 mm}
  \begin{tabularx}{\linewidth}{Y|YYYY}
    \toprule
         & PSNR       & SSIM \\
    \midrule
    SFFA & 34.82      & 0.9696 \\ 
    BFA  & \bf{34.95} & \bf{0.9703} \\      
    \bottomrule
  \end{tabularx}
  \label{tab:blur_map}
  \vspace{-2 mm}
\end{table}

\begin{table}[!t]
  \centering
  \caption{\textbf{Comparison of various token sparsity strategies.} The best results are highlighted in bold. The results are evaluated on the DVD dataset.}
  \vspace{-2 mm}
  \begin{tabularx}{\linewidth}{l|YYc}  
    \toprule
                    & PSNR       & SSIM        & GFLOPs \\
    \midrule
    Random 50\%     & 33.92      & 0.9651      & 133 \\
    100\%           & \bf{34.98} & \bf{0.9704} & 329  \\
    \midrule
    Top 25\%        & 34.78 & 0.9694 & \bf{127}  \\        
    Top 50\% (Ours) & 34.95      & 0.9703      & 133 \\
    \bottomrule
  \end{tabularx}
  \label{tab:selected_strategy}
  \vspace{-4 mm}
\end{table}

\subsection{Ablation Study}

\noindent \textbf{Effectiveness of BBFP.}
To evaluate the effectiveness of BBFP, we conduct an experiment by excluding BBFP from BSSTNet. 
As illustrated in Table~\ref{tab:structure}, the omission of BBFP in Exp. (b) results in a reduction of $0.21$ dB in PSNR and $0.0011$ in SSIM.
BFA plays an important role in preventing the introduction of blurry pixels from neighboring frames.
As shown in Table~\ref{tab:blur_map}, replacing BFA with Standard Flow-guided Feature Alignment results in a decline in performance.
To highlight the improved feature alignment capability of BBFP, we visualize the aligned features in Figure~\ref{fig:ab_bbfp}, comparing them with the standard feature bidirectional propagation (SFBP).
Benefiting from the incorporation of blur maps, BBFP prevents the propagation of blurry regions from the features of neighboring frames during the propagation process, resulting in sharper features.

\noindent \textbf{Effectiveness of BSST.}
To evaluate the effectiveness of BSST, we conduct an experiment by excluding BSST from BSSTNet. 
As shown in Table~\ref{tab:structure}, the omission of BSST in Exp. (c) results in a notable degradation of $0.85$ dB in PSNR and $0.0042$ in SSIM.
To further evaluate the effectiveness and efficiency of BSST, we compare different token sparsity strategies.
Table~\ref{tab:selected_strategy} demonstrates that using fewer token numbers or randomly selecting tokens will result in a significant decline in performance.
This result suggests that without guidance from the blur map, discarding tokens in the spatio-temporal domain results in the loss of valuable information in the video sequence.
Moreover, our sparsity strategy, which involves using the top 25\% of tokens, achieves performance comparable to using all tokens while utilizing only approximately \textbf{43\%} of the FLOPs.
This indicates that our sparsity strategy effectively leverages tokens in sharp regions within the video sequence.

\noindent \textbf{Comparison of Different Temporal Length.}
In Table~\ref{tab:temporal_length}, we present a comparison of the Standard Spatio-temporal Transformer (SST) under different sequence lengths in terms of PSNR, SSIM, Runtime, Memory, and GFLOPs. 
As the sequence length increases, the computational complexity of SST grows rapidly. 
In contrast, BSST's computational complexity is less affected by the sequence length, allowing BSST to utilize longer sequences and boost deblurring performance. 
Specifically, when the sequence length is 60, BSST shows a modest gain in PSNR and SSIM. 
Considering the balance between performance and computational load, we ultimately choose 48 as the length for the input video sequence. 

%% file: sec/5_conclusion.tex
\section{Conclusion}

In this paper, we present a novel approach for video deblurring, named BSSTNet.
Utilizing an understanding of the connection between pixel displacement and blurred regions in dynamic scenes, we introduce a non-learnable, parameter-free technique to estimate the blur map of video frames by employing optical flows.
By introducing Blur-aware Spatio-temporal Sparse Transformer (BSST) and Blur-aware Bidirectional Feature Propagation (BBFP), the proposed BSSTNet can leverage distant information from the video sequence and minimize the introduction of blurry pixels during bidirectional propagation.
Experimental results indicate that the proposed BSSTNet performs favorably against state-of-the-art methods on the GoPro and DVD datasets, while maintaining computational efficiency.